%% file: Spring_Gaus.tex
\documentclass[runningheads]{llncs}

\usepackage{eccv}
\usepackage{eccvabbrv}
\usepackage{graphicx}
\usepackage{booktabs}
\usepackage[accsupp]{axessibility}
\input{configs/packages.tex}

\usepackage{hyperref}
\usepackage[capitalize]{cleveref}
\input{configs/commands.tex}

\usepackage{orcidlink}

\begin{document}

\title{Reconstruction and Simulation of Elastic Objects with Spring-Mass 3D Gaussians}

\titlerunning{Spring-Gaus}

\author{Licheng Zhong\inst{1}$^*$ \and
        Hong-Xing Yu\inst{1} \and
        Jiajun Wu\inst{1} \and
        Yunzhu Li\inst{1,2,3}}

\authorrunning{L. Zhong et al.}
\institute{Stanford University, Stanford, USA \and
        Columbia University, New York, USA \and
        University of Illinois Urbana-Champaign, Champaign, USA \\
}

\maketitle

\def\thefootnote{*}\footnotetext{The work was done while L. Zhong was a visiting student at Stanford University. L. Zhong is now at Shanghai Jiao Tong University.}

\input{sec/0_abstract}

\input{sec/img/teaser.tex}

\input{sec/1_intro}

\input{sec/2_related}

\input{sec/4_method}

\input{sec/5_experiment}

\input{sec/7_conclusion}

%\newpage

\paragraph{\bf Acknowledgments.} The work is in part supported by the Amazon AICE Award, NSF RI \#2211258, \#2338203, ONR YIP N00014-24-1-2117, and ONR MURI N00014-22-1-2740.

\bibliographystyle{configs/splncs04}
\bibliography{egbib}

\clearpage
\appendix
\section*{Appendix}
\input{sec/supplementary}
\end{document}

%% file: configs/packages.tex
\usepackage{graphicx}
\usepackage{amsmath}
\usepackage{amssymb}
\usepackage{booktabs}
\usepackage{balance}
\usepackage{lipsum}
\usepackage{caption}
\usepackage{algorithm}
\usepackage{algorithmic}
\usepackage{enumerate}
\usepackage{engord}
\usepackage{bm}
\usepackage{bbm}
\usepackage{amstext}
\usepackage{comment}
\usepackage{color}
\usepackage{booktabs}
\usepackage{multirow}
\usepackage{mathtools}
\usepackage{setspace}
\usepackage{pifont}
\usepackage{dsfont}
\usepackage{cancel}
\usepackage[inline]{enumitem}
\usepackage[toc,page]{appendix}
\usepackage{hhline}

%% file: configs/commands.tex
\definecolor{Gray}{gray}{0.9}
\definecolor{White}{gray}{1}
\definecolor{WhiteGray}{rgb}{0.9, 0.9, 0.9}
\definecolor{DGray}{gray}{0.8}
\definecolor{DDDDGray}{gray}{0.3}
\definecolor{citecolor}{HTML}{0071bc}

\definecolor{DeltaColor}{rgb}{0.039,0.73,0.71}
\definecolor{SigmaColor}{rgb}{0.98,0.45,0.0}
\definecolor{AlphaColor}{rgb}{0,0,0.8}
\definecolor{BetaColor}{rgb}{0.8,0,0.8}
\definecolor{GammaColor}{rgb}{0.514,0.34,0.224}
\definecolor{EpsilonColor}{rgb}{0.353,0.725,0.906}
\definecolor{GreenColor}{rgb}{0.137,0.573,0.565}
\definecolor{RedColor}{rgb}{0.949,0.275, 0.224}
\definecolor{Cardinal}{rgb}{0.549,0.082,0.082}

\DeclareMathAlphabet\mathbfcal{OMS}{cmsy}{b}{n}

\newcommand{\colorRef}[1]{\textcolor{black}{#1}}
\crefname{figure}{\colorRef{Fig.}}{\colorRef{Figs.}}
\Crefname{figure}{\colorRef{Figure}}{\colorRef{Figures}}
\crefname{section}{\colorRef{Sec.}}{\colorRef{Secs.}}
\Crefname{section}{\colorRef{Section}}{\colorRef{Sections}}
\crefname{table}{\colorRef{Tab.}}{\colorRef{Tabs.}}
\Crefname{table}{\colorRef{Table}}{\colorRef{Tables}}
\Crefname{equation}{\colorRef{Eq.}}{\colorRef{Eqs.}}
\Crefname{equation}{\colorRef{Equation}}{\colorRef{Equation}}

\newcommand\method{Spring-Gaus\xspace}
\newcommand{\qheading}[1]{\noindent\mbox{\textbf{#1}\;}}

\newlength\savewidth\newcommand\shline{\noalign{\global\savewidth\arrayrulewidth
                \global\arrayrulewidth 1pt}\hline\noalign{\global\arrayrulewidth\savewidth}}

\newcommand{\projectURL}{\href{https://zlicheng.com/spring_gaus}{\tt{https://zlicheng.com/spring\_gaus}}}

%% file: sec/0_abstract.tex
\begin{abstract}
    Reconstructing and simulating elastic objects from visual observations is crucial for applications in computer vision and robotics. Existing methods, such as 3D Gaussians, model 3D appearance and geometry, but lack the ability to estimate physical properties for objects and simulate them. The core challenge lies in integrating an expressive yet efficient physical dynamics model. We propose \method, a 3D physical object representation for reconstructing and simulating elastic objects from videos of the object from multiple viewpoints. In particular, we develop and integrate a 3D Spring-Mass model into 3D Gaussian kernels, enabling the reconstruction of the visual appearance, shape, and physical dynamics of the object. Our approach enables future prediction and simulation under various initial states and environmental properties. We evaluate \method on both synthetic and real-world datasets, demonstrating accurate reconstruction and simulation of elastic objects. Project page: \projectURL.

    \keywords{System Identification \and Spring-Mass Model \and 3D Gaussians}
\end{abstract}

%% file: sec/img/teaser.tex
\begin{figure}[!h]
        \centering
        \includegraphics[width=\textwidth]{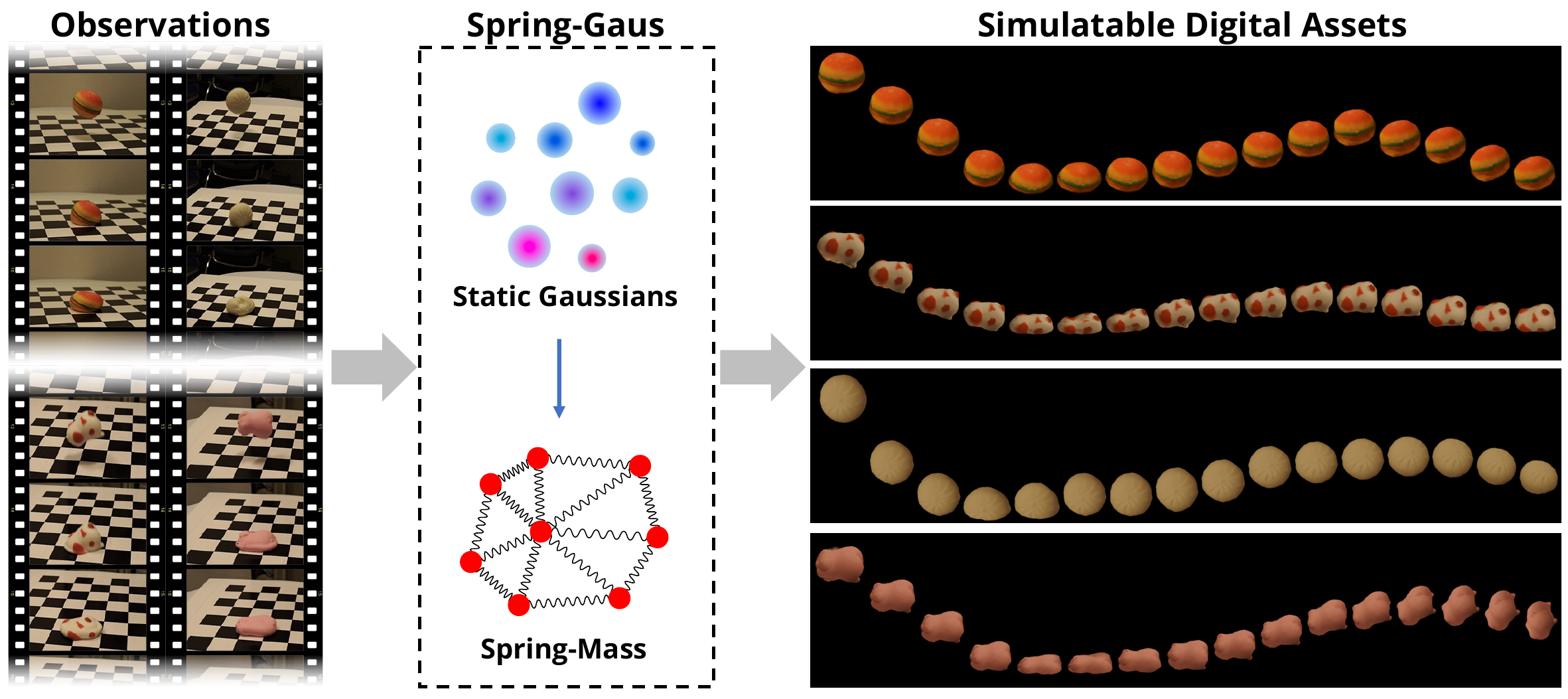}
        \caption{\small \method reconstructs the appearance, geometry, and physical dynamic properties of elastic objects from video observations. \method enables future predictions and simulations under different initial states and environmental conditions.}
        \label{fig:teaser}
\end{figure}

%% file: sec/1_intro.tex
\section{Introduction}
\label{sec:intro}

% Acquiring simulatable digital assets through visual perception plays a important role in computer vision and robotics, bridging the gap between digital and physical worlds with increasing interactivity and seamlessness. In VR/AR environments, these digital assets serve as the building blocks for creating immersive experiences that closely mimic real-world scenarios. For example, in medical training simulations, the precise modeling of human tissue and organ elasticity enables medical students to practice surgeries and procedures with remarkable realism in a safe, fully virtual environment. Similarly, in robotics, simulating object behaviors in virtual environments is essential for training and testing robotic systems. Robots that can perceive and interact with objects can undertake a broader range of tasks, including packing soft goods in warehouses, food preparation in kitchens, or providing assistance in homes.

Reconstructing and simulating elastic objects from visual observations poses a fundamental challenge in computer vision and robotics, with applications spanning virtual reality, augmented reality, and robotic manipulation. Accurately modeling the elasticity of objects is crucial for creating immersive experiences and enabling embodied agents to understand and interact with the elastic objects commonly encountered in our daily lives. However, accurately identifying the dynamics from vision still presents considerable challenges.
% to interact with a wider range of objects.

% Identifying digital assets from images or videos, however, poses significant challenges due to the complexity of objects and the necessity to capture spatial geometry and temporal dynamics. Recent advancements, such as D-NeRF \cite{pumarola2021dnerf}, an extension of Neural Radiance Fields (NeRF) \cite{mildenhall2020nerf}, have sought to model dynamic scenes through implicit representation. Conversely, approaches like Dynamic 3D Gaussians \cite{luiten2023dynamic}, 4D Gaussian Splatting \cite{wu20234dgaussians}, and Deformable 3D Gaussians \cite{yang2023deformable3dgs}, based on 3D Gaussian Splatting \cite{kerbl3Dgaussians}, propose modeling dynamic scenes via explicit representation using Gaussian kernels. Yet, these methods lack simulatability as they don't include physical properties. PhysGaussian \cite{xie2023physgaussian} offers a simulatable extension to 3D Gaussian Splatting but fails to capture objects' properties from observations due to its non-optimizable pipeline. PAC-NeRF \cite{li2023pacnerf} integrates physics by combining DirectVoxGO \cite{SunSC2022dvgo} with Material Point Method (MPM) \cite{stomakhin2012mpm,hu2018mlsmpm} simulations to infer physical properties from multi-view imagery, yet optimizing physical parameters for each point remains challenging, particularly for heterogeneous objects. Moreover, its implicit representation can degrade the authenticity of appearance over time.

Existing methods for dynamic scene reconstruction, such as 3D Gaussians and their dynamic extensions~\cite{kerbl3Dgaussians,luiten2023dynamic,wu20234dgaussians,yang2023deformable3dgs}, have made significant progress in capturing the temporal changes of the appearance and geometry of objects. However, these methods do not capture the physical properties of the reconstructed objects; thus, they typically cannot predict the future dynamics of these objects.
While a few recent approaches using MPM~\cite{stomakhin2012mpm,hu2018mlsmpm} have attempted to integrate physics-based priors into 3D object representations, such as PAC-NeRF~\cite{li2023pacnerf}. Their ability to handle real, especially heterogeneous, objects is limited, as they assume a known material model and only assigns a global physical parameter to the entire object which restricts its adaptability to real objects.
Assigning learnable physical parameters to each particle in an MPM is theoretically possible. However, in practice, it incurs extreme computational costs since MPM requires tens of thousands of dense points. In addition, PAC-NeRF~\cite{li2023pacnerf} use a implicit grid representation, due to the computational demands, the resolution of the grid is limited, which can lead to a loss of detail in the appearance modeling when using real or noisy data.
% The core challenge is that some physics simulators, such as Material Point Method (MPM) \cite{stomakhin2012mpm,hu2018mlsmpm} used in PAC-NeRF, have some physical assumptions and consider a more restrictive model class that lead to differences from real data and difficulty in generalizing on data obtained by other simulation methods. 
Thus, the core challenge in reconstructing and predicting object dynamics lies in developing and integrating an \emph{expressive} and \emph{efficient} physical model for the dynamics. The physical dynamics model should be expressive enough to capture the motions of elastic objects, including collisions, deformations, and bouncing. It must also be efficient and conducive to inverse parameter estimation through gradient-based optimization.

In this work, we propose \method, a 3D object representation that integrates a 3D Spring-Mass model. Our model represents the elastic object dynamics properties through a learnable system of mass points and springs. Our design is expressive in that it assumes a general and widely applicable physical model class. With a learnable topology and physical parameters, \method can model complex deformation and motion for heterogeneous elastic objects. In addition to expressiveness, \method is also highly efficient for the inverse optimization of physical parameters due to its differentiable nature.
% Moreover, Our 3D \method can be adopted on a sparse point cloud distribution which is more flexible and better suited of gradient-based optimization. 

\method enables reconstructing and simulating elastic objects from sparse multi-view videos (\cref{fig:teaser}).
To overcome the intrinsic difficulty in optimization, we propose a reconstruction pipeline that decouples physical parameter reconstruction from appearance and geometry reconstruction.  \method requires only a few multi-view videos for physical parameter identification and is robust to the quality of geometry reconstruction.

We evaluate the effectiveness of \method on both synthetic and real-world datasets, demonstrating its ability to accurately reconstruct and simulate elastic objects. Our \method allows accurate future prediction and simulations under varying initial states and environmental parameters, showcasing its potential for applications in predictive visual perception and immersive experiences. In summary, the main contributions of this work are threefold:
\begin{itemize}
    \item We propose \method, which incorporates an expressive yet efficient 3D Spring-Mass model for reconstructing and simulating elastic objects.
    \item We introduce a pipeline to reconstruct \method from multi-view videos of the object. We decouple the appearance and geometry reconstruction from the physical dynamics reconstruction for more effective optimization.
    \item We demonstrate the effectiveness of \method on both synthetic and real-world datasets, showcasing accurate reconstruction and simulation of elastic objects. This includes capabilities for future prediction and simulation under varying initial configurations.
\end{itemize}

%% file: sec/2_related.tex
\section{Related Work}

\qheading{3D Object Representations:}
Traditionally, 3D objects are often represented by point clouds, meshes, and voxels. Recently, neural 3D object representation has become popular due to the efficiency and flexibility. For example,
Scene Representation Networks (SRNs) \cite{sitzmann2019srns} and DeepSDF~\cite{park2019deepsdf} represent significant advancements in 3D scene and object representation, treating 3D objects as continuous functions that map world coordinates to a feature representation of local object properties. Over the past few years, NeRF \cite{mildenhall2020nerf} and its successors \cite{yu2020pixelnerf, mvsnerf, barron2021mipnerf, liu2022neuray, xu2022point} have demonstrated the efficacy of neural networks in capturing continuous 3D scenes and objects through implicit representation. DirectVoxGO \cite{SunSC2022dvgo} accelerates NeRF's approach by substituting the MLP with a voxel grid. Furthermore, 3D Gaussian Splatting \cite{kerbl3Dgaussians} has emerged as a method for real-time differentiable rendering, representing scenes and objects with 3D Gaussians. Extending this approach, DreamGaussian \cite{tang2023dreamgaussian} applies 3D Gaussians for 3D object generation. Unlike these approaches, our method focuses on reconstructing simulatable 3D objects. Recently, PhysGaussian \cite{xie2023physgaussian} integrated physical simulation into 3D Gaussians using a customized Material Point Method, allowing forward simulation of reconstructed objects. In contrast to PhysGaussian, our work focuses on system identification from raw videos and supports both the forward simulation and the inverse reconstruction of physical objects.

\qheading{Dynamic Scene Reconstruction:} The modeling of dynamic scenes has seen significant progress with the adoption of NeRF \cite{pumarola2021dnerf,li2023pacnerf,li2021nsff,tretschk2021nonrigid,park2021nerfies,guan2022neurofluid,li20223d,driess2023learning,dynibarli2023,flowwang2023,hypernerfpark2021} and 3D Gaussian \cite{luiten2023dynamic,yang2023deformable3dgs,wu20234dgaussians,lin2023gaussian,yu2023cogs,huang2024sc,kratimenos2024dynmf,yang2023gs4d} representations. D-NeRF \cite{pumarola2021dnerf} introduces an extension to NeRF, capable of modeling dynamic scenes from monocular views by optimizing an underlying deformable volumetric function. Furthermore, Dynamic 3D Gaussians \cite{luiten2023dynamic} focus on optimizing the motion of Gaussian kernels for each frame, presenting an efficient method to capture scene dynamics. Deformable 3D Gaussians \cite{yang2023deformable3dgs} propose a novel approach by learning Gaussian distributions in a canonical space, complemented with a deformation field for modeling monocular dynamic scenes. Meanwhile, 4D Gaussian Splatting \cite{wu20234dgaussians} introduces a hybrid model combining 3D Gaussians with 4D neural voxels. PAC-NeRF \cite{li2023pacnerf} delves into the integration of Lagrangian particle simulation with Eulerian scene representation, exploring a new aspect of scene dynamics.
Among them, PAC-NeRF \cite{li2023pacnerf} is the most relevant to our work. PAC-NeRF learns physical parameters used in the Material Point Method (MPM) for better dynamic reconstruction and prediction. However, PAC-NeRF assumes known material models that restrict its adaptability and applicability to complex real objects. In contrast, our work integrates a 3D Spring-Mass model that is both expressive and efficient, allowing reconstructing and simulating real elastic objects.

\qheading{Physics-Informed Learning:} Physics-Informed Learning has emerged as a prominent research direction since the introduction of Physics-Informed Neural Networks (PINNs) \cite{PINN}. Li \etal~\cite{li2018learning} and Sanchez-Gonzalez \etal \cite{sanchezgonzalez2020learning} introduced Graph Network-based simulators within a machine learning framework. INSR-PDE \cite{chenwu2023insr-pde} tackles time-dependent partial differential equations (PDEs) using implicit neural spatial representations. NCLaw \cite{ma2023nclaw} focuses on learning neural constitutive laws for PDE dynamics. DiffPD \cite{du2021_diffpd} presents a differentiable soft-body simulator. Li \etal \cite{li2023learnpde} have made strides in learning preconditioners for conjugate gradient PDE solvers. Chu \etal \cite{chu2022moke} and Yu \etal~\cite{yu2024inferring} modeled smoke in neural density and velocity fields. Neural Flow Maps \cite{deng2023neural} integrate fluid simulation with neural implicit representations. Deng \etal \cite{deng2023vortex} introduced a novel differentiable vortex particle method for fluid dynamics inference. DANO \cite{le2023danos} is the most relevant work to ours. In particular, DANO develops a differentiable simulation for rigid objects represented by NeRFs, allowing reconstruction and simulation from a few videos. In contrast, our work focuses on elastic objects, which involve a fundamentally new set of challenges and require a redesign of the physical models.

%% file: sec/4_method.tex
\section{Approach}

We propose \method that integrates a 3D Spring-Mass model with 3D Gaussians for elastic object reconstruction and simulation.
Specifically, we are given a set of $O$ calibrated videos from multiple viewpoints of an elastic object in motion, denoted as
% $\{{I_{i,j}}\}_{i=0,j=0}^{n,f}$
$ \{{I^{o, f}}\}_{o=1,f=1}^{O,F}$,
where $F$ is the number of video frames and $I^{o, f}$ denotes a 2D image at video $o$ and frame $f$. Our goal is to reconstruct the appearance, geometry, and physical dynamics parameters of \method.

In the following, we first introduce the representation for the static properties, i.e., appearance and geometry, and then the dynamics properties, i.e., the 3D Spring-Mass model. Then we describe our reconstruction pipeline.
We show an overview of our reconstruction pipeline in \cref{fig:pipeline}.

\subsection{Appearance and Geometry Representation}

% Given a set of posed videos captured by $n$ cameras over $f$ frames, denoted as $\{{I_{i,j}}\}_{i=0,j=0}^{n,f}$, we utilize the first frame from each sequence, $\{I_{i,0}\}_{i=0}^{n}$, to reconstruct the initial static appearance and geometry.
% %
Following 3D Gaussians Splatting \cite{kerbl3Dgaussians}, we use a set of 3D Gaussian kernels to represent the appearance and shape of an object. These kernels are described by a full 3D covariance matrix $\Sigma$, defined in world space and centered at point $\boldsymbol{\mu}$:
\begin{equation}
    G(\boldsymbol{\tau}) = e^{-\frac{1}{2}(\boldsymbol{\tau})^T\Sigma^{-1}(\boldsymbol{\tau})},
\end{equation}
where $\boldsymbol{\mu}$ is the mean of the Gaussian distribution, and $\boldsymbol{\tau}$ is the independent variable of the Gaussian distribution.
In our formulation, we assume all Gaussian kernels are isotropic, the covariance matrix $\Sigma$ can be controlled by a scalar $s \in \mathbb{R}$.

The splatting process, designed to render Gaussian kernels into a 2D image, involves two main steps: (1) projecting the Gaussian kernels into camera space, and (2) rendering the projected kernels into image space. The projection process is defined as:
\begin{equation}
    \Sigma^{\prime} = JW\Sigma W^TJ^T,
\end{equation}
where $\Sigma^{\prime}$ represents the covariance matrix in camera coordinates, $W$ is the transformation matrix, and $J$ denotes the Jacobian of the affine approximation of the projective transformation. The color $C$ of each pixel is rendered as:
\begin{equation}
    C = \sum_{i\in N}T_i\alpha_i \boldsymbol{c}_i,
    \label{eq:rendering}
\end{equation}
where $N$ is the total number of kernels and $T_i$ represents the transmittance, defined as $\Pi_{j=1}^{i-1}\alpha_j$. The term $\alpha_i$ denotes the alpha value for each Gaussian, which is calculated using the expression $1-e^{-\sigma_i\delta_i}$, where \(\sigma_i\) is the opacity factor. Additionally, $\boldsymbol{c}_i$ refers to the color of the Gaussian along the ray within the interval $\delta_i$.

For each kernel, the learnable parameters include a point center $\boldsymbol{\mu}$, a scaling scalar $s$, a color vector $\boldsymbol{c}$, and an opacity value $\sigma$.

\input{sec/img/pipeline.tex}

\subsection{Physics-Based Dynamics with 3D Spring-Mass Model}

We aim to develop and integrate an \emph{expressive} yet \emph{efficient} physics-based dynamics model. With these two design goals in mind, we introduce the 3D Spring-Mass model which represents the elastic object dynamics by a learnable spring-mass system. Our \method does not assume any material type, and it is expressive to model real heterogeneous elastic objects. Besides being expressive, our dynamics model should also be efficient and amenable to gradient-based optimization for parameter estimation.

Recall that the appearance and geometry representations include a set of Gaussian kernels $\{G_i\}_{i=1}^{N}$, parameterized by point centers $\boldsymbol{X}=\{\boldsymbol{\mu}_i\}_{i=1}^{N}$, scaling scalars $\{s_i\}_{i=1}^{N}$, color vectors $\{\boldsymbol{c}_i\}_{i=1}^{N}$, and opacity values $\{\sigma_i\}_{i=1}^{N}$.
To manage complexity for efficient simulation, we introduce volume sampling to generate a set of anchor points $\boldsymbol{A}=\{\boldsymbol{x}_i\}_{i=1}^{N_A}$ (each $\boldsymbol{x}$ represents a spatial point), defined by:
\begin{equation}
    \boldsymbol{A}=\{\boldsymbol{x}_i\}_{i=1}^{N_A} = \mathcal{V}(\boldsymbol{X}),
\end{equation}
where $N_A$ denotes the number of anchors, and $\mathcal{V}$ denotes the sampling function.

Our 3D Spring-Mass physical model can simulate the motion of anchor points $\boldsymbol{A}$, assuming each anchor has a mass $m_i$ and an initial velocity $\boldsymbol{v}_i$. Each anchor $\boldsymbol{x}_i$ connects to its $n_k$ nearest neighbors $\boldsymbol{N}_i = \{\boldsymbol{x}_{i,j}\}_{j=1}^{n_k}$ through springs $\boldsymbol{L}$:
\begin{equation}
    \boldsymbol{L} = \{l_{i,j}\}_{i=1,j=1}^{N_A,n_k} = \mathrm{knn}(\boldsymbol{A}, \boldsymbol{A}, n_k),
\end{equation}
where $l_{i,j}$ denotes the spring's length between $\boldsymbol{x}_i$  and $\boldsymbol{x}_{i,j}$, and $\mathrm{knn}$ denotes the k-nearest neighbors function. Each spring is characterized by a stiffness $k_{i,j}$ and a damping factor $\zeta_{i,j}$.

To update the positions of the kernels, we first measure the distance between each kernel center and its $n_b$ nearest anchors at the dynamic simulation's onset:
\begin{equation}
    \{d_{i,j}\}_{i=1,j=1}^{N,n_b} = \mathrm{knn}(\boldsymbol{X}, \boldsymbol{A}, n_b).
\end{equation}
For each timestep $t$, the forces $\boldsymbol{F}_i^{t}$  acting on each anchor point $\boldsymbol{x}_i^{t}$ are calculated as follows:
\begin{equation}
    \boldsymbol{F}_i^{t} = {\boldsymbol{F_{k}}}_{i}^{t} + {\boldsymbol{F_{\zeta}}}_{i}^{t} + m_i \boldsymbol{g},
\end{equation}
where ${\boldsymbol{F_{k}}}_{i}^{t}$, ${\boldsymbol{F_{\zeta}}}_{i}^{t}$, and $\boldsymbol{g}$ represent the spring force, the damping force, and gravitational acceleration, respectively.

Then, for each spring $\boldsymbol{L}_{i,j}$, the spring force ${\boldsymbol{F_{k}}}_{i,j}^{t}$ and damping force ${\boldsymbol{F_{\zeta}}}_{i,j}^{t}$ are determined by:
\begin{gather}
    {\boldsymbol{F_{k}}}_{i,j}^{t} = -\eta_j \cdot k_{i,j} \left( \left\| \boldsymbol{x}_i^{t} - \boldsymbol{x}_{i,j}^{t} \right\| - l_{i,j} \right) \frac{\boldsymbol{x}_i^{t} - \boldsymbol{x}_{i,j}^{t}}{\left\| \boldsymbol{x}_i^{t} - \boldsymbol{x}_{i,j}^{t} \right\|} \cdot {\Big\vert \left\| \boldsymbol{x}_i^{t} - \boldsymbol{x}_{i,j}^{t} \right\| - l_{i,j} \Big\vert}^{p_k},
    \label{eq:spring_force} \\
    {\boldsymbol{F_{\zeta}}}_{i,j}^{t} = \left(-\zeta_{i,j} \left( \boldsymbol{v}_i^{t} - \boldsymbol{v}_{i,j}^{t} \right) \frac{\boldsymbol{x}_i^{t} - \boldsymbol{x}_{i,j}^{t}}{\left\| \boldsymbol{x}_i^{t} - \boldsymbol{x}_{i,j}^{t} \right\|} \right) \cdot \frac{\boldsymbol{x}_i^{t} - \boldsymbol{x}_{i,j}^{t}}{\left\| \boldsymbol{x}_i^{t} - \boldsymbol{x}_{i,j}^{t} \right\|},
\end{gather}
where $\eta$ is a soft vector which will be discussed later and $p_k$ is a hyperparameter that determines the nonlinearity of the spring force. When $p_k$ is set to $0$, \cref{eq:spring_force} becomes Hooke's law, and for positive values of $p_k$, the spring force becomes a nonlinear function of the spring's length.
The cumulative forces acting on each anchor point $\boldsymbol{x}_i^{t}$  are expressed as:
\begin{equation}
    \boldsymbol{F}_i^{t} = \sum_{j=1}^{n_k}{\boldsymbol{F_{k}}}_{i,j}^{t} + \sum_{j=1}^{n_k}{\boldsymbol{F_{\zeta}}}_{i,j}^{t} + m_i \boldsymbol{g}.
\end{equation}
Anchor points $\boldsymbol{A}$'s positions and velocities are updated using semi-implicit Euler integration:
\begin{gather}
    \hat{\boldsymbol{v}}_i^{t+1} = \boldsymbol{v}_i^{t} + \frac{\boldsymbol{F}_i^{t}}{m_i} \Delta t,
    \label{eq:update_rule0}
    \\
    \hat{\boldsymbol{x}}_i^{t+1} = \boldsymbol{x}_i^{t} + \boldsymbol{v}_i^{t+1} \Delta t,
    \label{eq:update_rule1}
\end{gather}
and a boundary condition $\mathcal{B}$ is applied to the anchor points $\boldsymbol{A}$ to model the interactions with the environment:
\begin{equation}
    \boldsymbol{x}_i^{t+1}, \boldsymbol{v}_i^{t+1} = \mathcal{B}(\hat{\boldsymbol{x}}_i^{t+1}, \hat{\boldsymbol{v}}_i^{t+1}).
\end{equation}
The position of each Gaussian kernel $\boldsymbol{\mu}_i$ is updated through Inverse Distance Weighting (IDW) interpolation to reflect the dynamic changes accurately:
\begin{equation}
    \boldsymbol{\mu}_i^{t+1} = \frac{\sum_{j=1}^{n_b}{\boldsymbol{x}_{i,j}^{t+1}} \cdot (1/{(d_{i,j})}^{p_b})}{\sum_{j=1}^{n_b}(1/{(d_{i,j})}^{p_b})},
    \label{eq:idw}
\end{equation}
where $p_b$ is a positive real number that determines the diminishing influence of anchor points with distance. To render the image $\hat{I}$ at a specific camera and frame, we use the updated positions of the Gaussian kernels following the rendering equation \cref{eq:rendering}.

\qheading{Soft Vector for Springs Connection:}
In our formulation above, the $n_k$ is a hyperparameter which is the number of connected springs for each anchor. However, the choice of $n_k$ will directly affect the simulation results markedly. The bigger value of $n_k$, the object will behave more rigidly, and a smaller value of $n_k$ leads to a noticeably softer behavior of the point cloud.
To address the significant impact that the value of $n_k$ has on the simulation, we introduce a mitigation strategy by applying a soft vector $\eta = [\eta_0,\eta_1,...,\eta_{n_k}]$. This vector controlled by a learnable parameter $\kappa$ (shared by all anchors) is used to modulate the number of connected springs, thereby adjusting the system's response to a different object. Given an empirical value $n_c$, the soft vector $\eta$ is calculated as:
\begin{equation}
    \eta_j = \left\{
    \begin{aligned}
         & 1                                                             &  & j \leq n_c,       \\
         & \mathrm{clamp}(2-\exp(\mathrm{softplus}(\kappa))^{j-n_c},0,1) &  & n_c < j \leq n_k.
    \end{aligned}
    \right.
\end{equation}

\subsection{Optimization}

To allow efficient optimization, we simplify our model by reducing the number of learnable parameters without changing the essential expressiveness. In our simplified approach, we standardize the mass of every anchor to a constant value $m_0$, and control all damping factors using a singular parameter $\zeta_0$. These two parameters are fixed, eliminating their variability from the optimization process. Furthermore, we introduce a unified parameter $k_i$ for each anchor $\boldsymbol{x}_i$ to control the spring stiffness of the springs connected to it, simplifying the model without compromising its functional integrity. The spring stiffness $k_{i,j}$ and damping factor $\zeta_{i,j}$ are thus given as follows:
\begin{gather}
    k_{i,j} = k_i / l_{i,j}, \\
    \zeta_{i,j} = \zeta_0 / l_{i,j}.
    \label{eq:stiffness_define}
\end{gather}

Note that, in \cref{eq:stiffness_define} stiffness is defined at anchor points rather than on springs themselves to simplify the optimization\textemdash we only need to optimize $N_A$ stiffness coefficients instead of all the springs, which is on the order of $n_k\cdot N_A$. By this simplification, our model maintains computational efficiency and ease of optimization, while still capturing the essential dynamics of the object.

Summarized, the learnable parameters in our model now include:
\begin{itemize}
    \item $\boldsymbol{v}_0$: the initial velocity vector, providing a baseline movement pattern for the simulation;
    \item $\{{k_i}\}_{i=1}^{N_A}$: the individual stiffness parameters for each anchor, allowing for localized adjustments to spring stiffness;
    \item $\kappa$: a parameter governing the modulation of the soft vector, facilitating fine-tuned control over the spring dynamics;
    \item $\Theta(\mathcal{B(\cdot)})$: parameters defining the boundary conditions ($\Theta$ is learnable parameters), such as the friction coefficient, which influence the simulation's physical realism.
\end{itemize}

There exist $n_t$ timesteps between each of the two keyframes. At each timestep, $\boldsymbol{x}^t$ is computed from $\boldsymbol{x}^{t-1}$ using the update rule \cref{eq:update_rule0,eq:update_rule1}. We only optimize physical parameters at each keyframe (when $t = 0, n_t, 2n_t, 3n_t, \ldots$) based on the visual observations until that time. We increase the value of $n_t$ as the parameters converge. This approach balances identification accuracy and computational demand.

Same with 3D Gaussian Splatting \cite{kerbl3Dgaussians}, we define our loss function as a weighted combination of the $\mathcal{L}_1$ norm and the Structural Similarity Index Measure (D-SSIM) $\mathcal{L}_{\text{d-ssim}}$, applied between the input images $I$ and the rendered images $\hat{I}$. This is formally expressed as:

\begin{equation}
    \mathcal{L} = (1-\lambda_{\text{d-ssim}})\mathcal{L}_{1} + \lambda_{\text{d-ssim}}\mathcal{L}_\text{d-ssim},
\end{equation}
where $\lambda_{\text{d-ssim}}$ is a weighting coefficient that balances the contribution of the $\mathcal{L}_1$ norm and the D-SSIM term to the overall loss. We decouple our optimization into a few stages, as illustrated in \cref{fig:pipeline}.

\qheading{Static Reconstruction:} Our optimization starts by taking the first frames of the multi-view videos and using them to reconstruct the appearance and geometry of the object.

\qheading{3D Gaussians Refinement:} During dynamic simulation of the anchor points, the 3D Gaussian centers $\boldsymbol{X}_0=\{\boldsymbol{\mu}^0_i\}_{i=1}^{N}$ are computed by IDW interpolation outlined in \cref{eq:idw}. This leads to a slight appearance drift from the static reconstruction.
Therefore, we refine the parameters of the Gaussian kernels\textemdash scaling scalar $s$, color vector $\boldsymbol{c}$, and opacity value $\sigma$, excluding points center $\boldsymbol{\mu}$\textemdash also at the first frame, before the dynamic simulation.
Since the Gaussian kernels' position from IDW interpolation is slightly different from the static reconstruction results, this refinement process enables the Gaussian kernels with the interpolated spatial position to render the correct object appearance.
% This refinement ensures the Gaussian kernels adhere to the IDW interpolation results.

\qheading{Dynamic Reconstruction:}
Since our efficient dynamics simulation is fully differentiable with respect to the learnable parameters, we can optimize the physical parameters through differentiable simulation and differentiable rendering with our loss function $\mathcal{L}$.

%% file: sec/img/pipeline.tex
\begin{figure}[!t]
        \begin{center}
                \includegraphics[width=\linewidth]{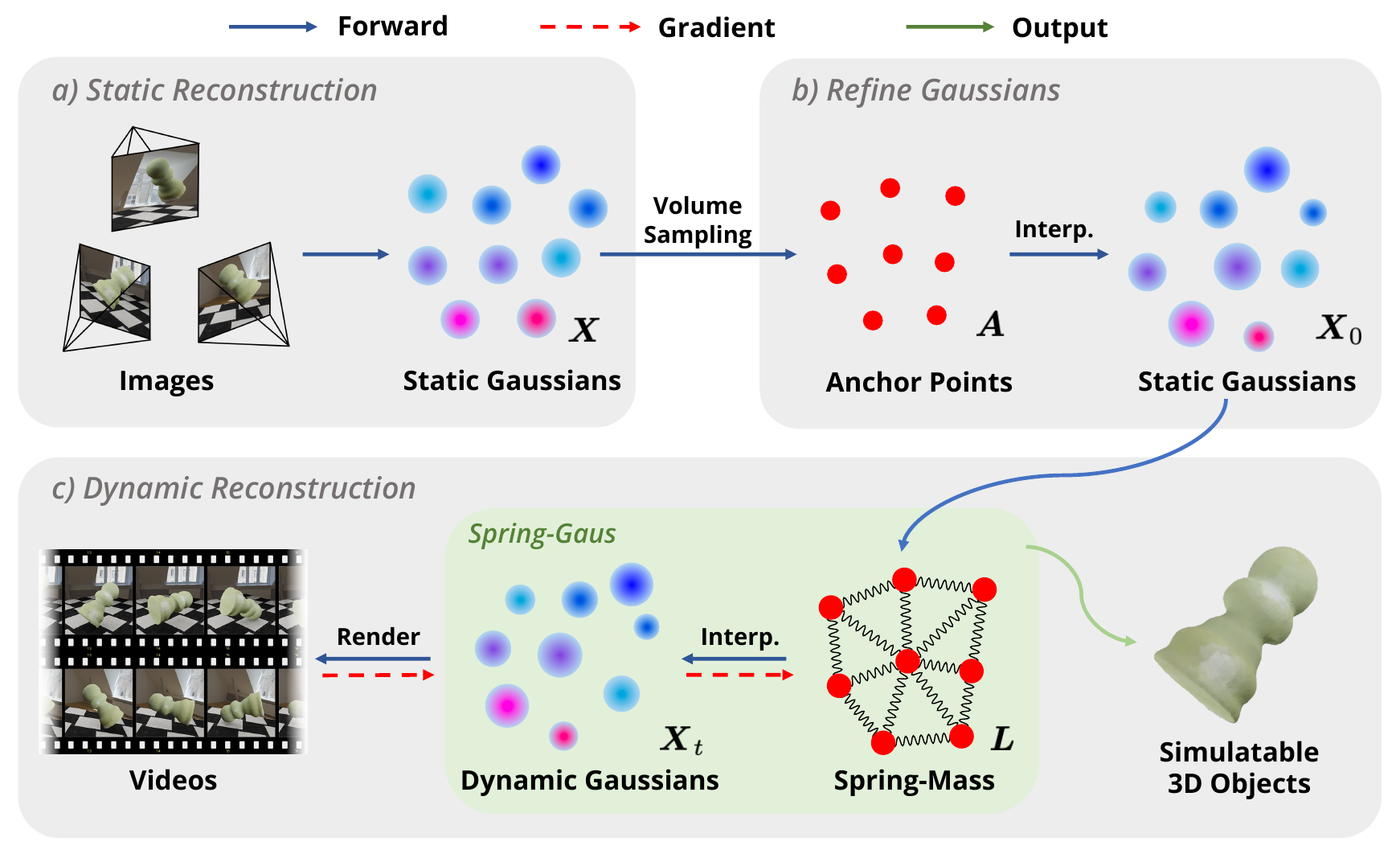}
        \end{center}
        \caption{\small \textbf{Overview of \method reconstruction pipeline:} \textbf{(a) Static Scene Reconstruction:} We start by reconstructing static 3D Gaussians from the first frames of the multiview videos. \textbf{(b) Refining 3D Gaussians:} We extract a set of anchor points to allow efficient simulation, which leads to appearance drift. We refine the 3D Gaussians to better model the appearance during simulation. \textbf{(c) Dynamic Reconstruction:} Our 3D Spring-Mass model simulates anchor points and updates the positions of Gaussian kernels. Upon completion of optimization, we obtain a simulatable 3D object that accurately models its dynamics.}
        \label{fig:pipeline}
\end{figure}

%% file: sec/5_experiment.tex
\section{Experiment}

\input{sec/exp/dataset.tex}

\input{sec/exp/detail.tex}

\input{sec/img/qualitative_mpm.tex}

\input{sec/exp/qualy_quanty.tex}

\input{sec/exp/real_sample.tex}

\input{sec/exp/digital.tex}

%% file: sec/exp/dataset.tex
\subsection{Datasets}

\qheading{Synthetic Data.} We evaluate \method using a synthetic dataset. We first collected fourteen 3D models, including point clouds and meshes, to use as initial point clouds for our simulations. Some of them are sourced from PAC-NeRF~\cite{li2023pacnerf} and OmniObject3D~\cite{wu2023omniobject3d}. Following PAC-NeRF, we employ the Material Point Method~\cite{stomakhin2012mpm,hu2018mlsmpm} to simulate the dynamics of elastic objects to generate synthetic data. Our dataset features elastic objects with various stiffness levels and diverse geometric forms. Multi-view RGB videos are rendered using Blender~\cite{blender}, with each sequence comprising 10 viewpoints and 30 frames at a resolution of 512$\times$512. The initial 20 frames are utilized for dynamic reconstruction, while the subsequent 10 frames are dedicated to evaluating future prediction performance.

\qheading{Real-World Data.} In addition to synthetic datasets, we further assess \method using captured real-world examples. Our collection process distinguishes between static scenes and dynamic multi-view videos. For static scenes, we position each object on a table and capture 50-70 images from various viewpoints within the upper hemisphere surrounding the object. To obtain camera poses for static scenes, we utilize the off-the-shelf Structure-from-Motion toolkit COLMAP \cite{schoenberger2016sfm, schoenberger2016mvs}. The dynamic aspect of our dataset is represented through multi-view RGB videos, recorded from three distinct viewpoints, at a resolution of 1920$\times$1080. The camera parameters for dynamic scenes are obtained through calibration using a checkerboard. SAM~\cite{kirillov2023segany} is used to obtain object masks.

%% file: sec/exp/detail.tex
\subsection{Implementation Details}
\label{sec:details}

\input{sec/img/regist.tex}

When learning the 3D Gaussians, the distribution of Gaussian kernels is highly dependent on the initial points. Gaussian kernels tend to concentrate on the surface if the initial points are derived from SfM results. However, we initialize a large number of points inside a cube, resulting in a more uniform distribution of the final kernels.

For real-world samples, we collect static and dynamic scenes separately because three viewpoints are insufficient for effective 3D Gaussian reconstruction~\cite{kerbl3Dgaussians}. This results in the Gaussian kernels for static and dynamic scenes being in different coordinate systems. Therefore, for real samples, we employ a registration network before dynamic reconstruction to align the Gaussian kernels from static scene coordinates to dynamic scene coordinates, as shown in \cref{fig:regist}. Specifically, we optimize a scale factor $s_r$, a translation vector $\boldsymbol{t}_r$, and a rotation vector $\boldsymbol{r}_r$ for the registration. We represent 3D rotations using a continuous 6D vector $\boldsymbol{r}_r \in \mathbb{R}^6$, which has been shown to be more amenable for gradient-based optimization~\cite{Zhou_2019_6dpose}.
However, slight deformations of objects and variations in lighting conditions and exposure times during data capture are inevitable. These variations can result in color discrepancies between frames across time and viewpoints. The color discrepancy is the major noise source that hinders registration and reconstruction. Thus, we refine our approach by computing the loss function between mask images, incorporating both mask center loss and perceptual loss into our model. Consequently, the revised loss function for real samples is expressed as:

\begin{equation}
    \mathcal{L} = (1-\lambda_{\text{d-ssim}})\mathcal{L}_{1} + \lambda_{\text{d-ssim}}\mathcal{L}_\text{d-ssim} + \lambda_{\text{center}}\mathcal{L}_\text{center} + \lambda_{\text{percep}}\mathcal{L}_\text{percep},
\end{equation}
where $\lambda_{\text{d-ssim}}=0.8$, $\lambda_{\text{center}}=1.0$, and $\lambda_{\text{percep}}=0.1$ are the weighting coefficients. Here, $\mathcal{L}_\text{center}$ quantifies the discrepancy between the center coordinates of the rendered and ground truth images, while $\mathcal{L}_\text{percep}$ represents the perceptual loss~\cite{johnson2016perceptual}, which is based on the VGG16 architecture~\cite{simonyan2014vgg}.

%% file: sec/img/regist.tex
\begin{figure}[!t]
        \begin{center}
                \includegraphics[width=0.8\linewidth]{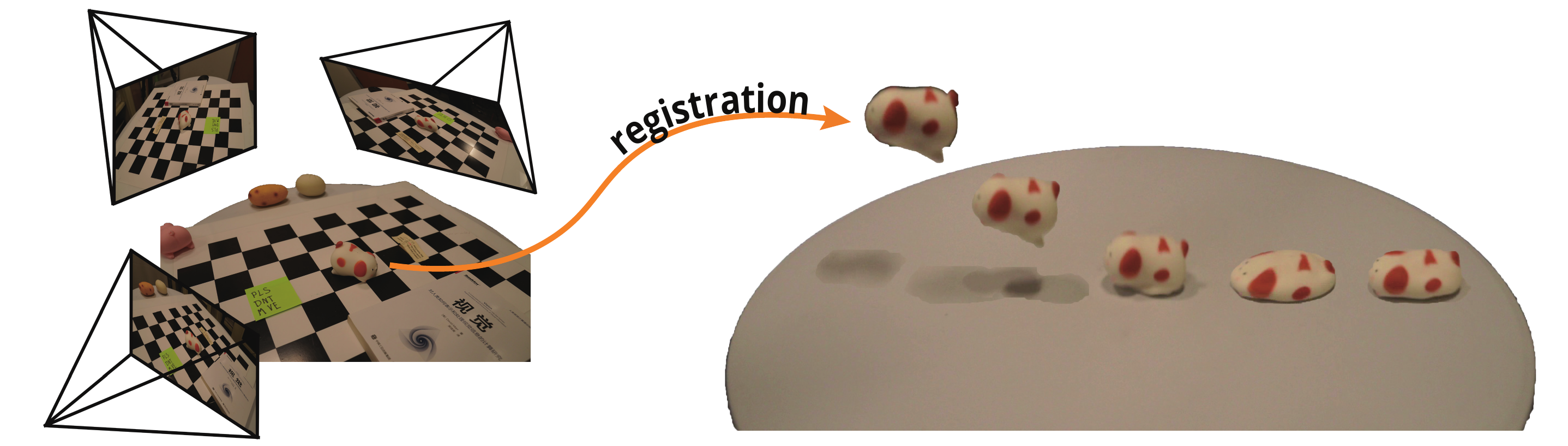}
        \end{center}
        \caption{\small Registration from static scene to dynamic scene for real-world sample.}
        \label{fig:regist}
\end{figure}

%% file: sec/img/qualitative_mpm.tex
\begin{figure}[!t]
        \begin{center}
                \includegraphics[width=\linewidth]{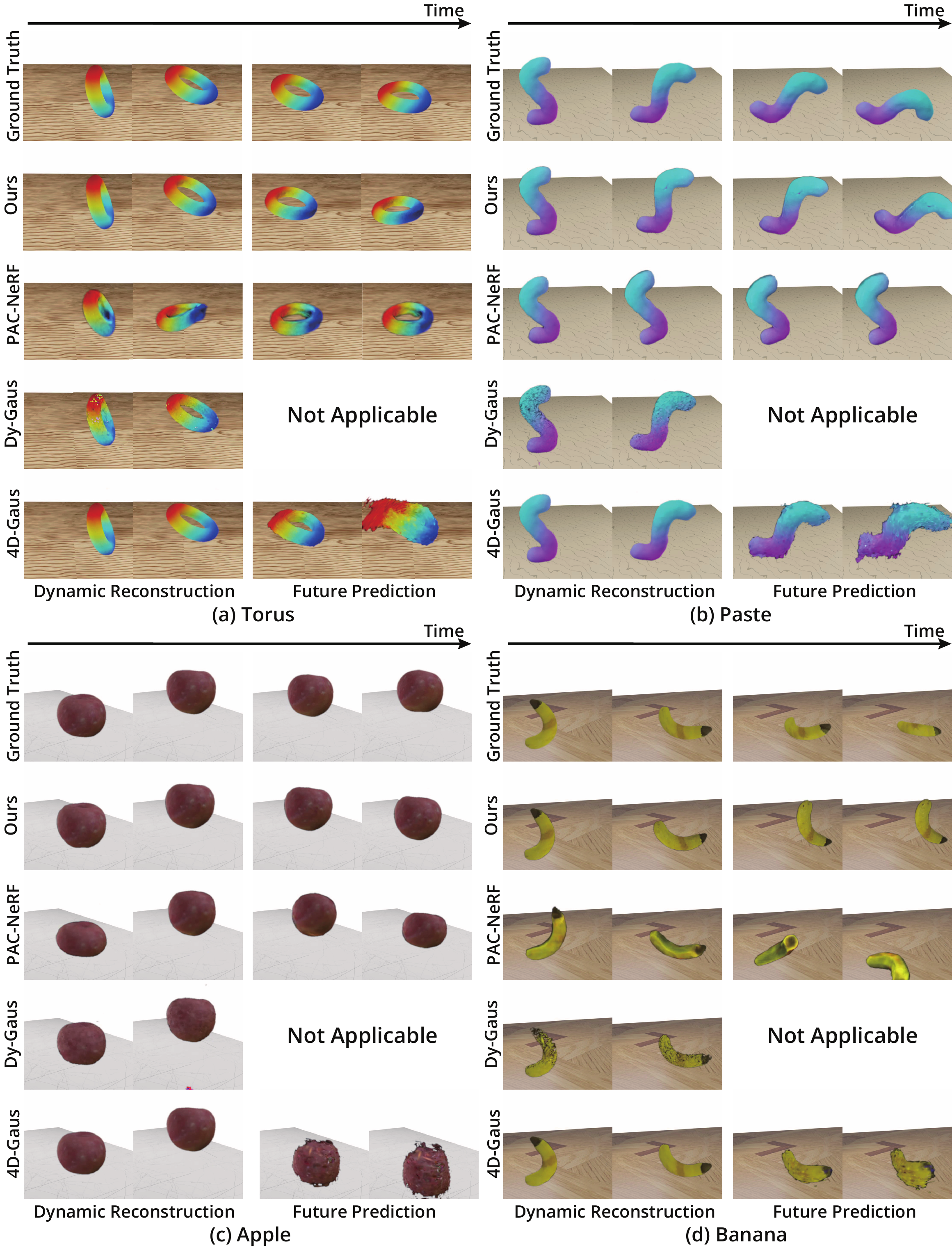}
        \end{center}
        \caption{\small \textbf{Qualitative results on synthetic data.} Compared with PAC-NeRF \cite{li2023pacnerf}, Dynamic 3D Gaussians \cite{luiten2023dynamic} and 4D Gaussian Splatting \cite{wu20234dgaussians}, \method can maintain a good geometry and appearance while reconstructing reasonable dynamics.}
        \label{fig:qualitative_mpm}
\end{figure}

%% file: sec/exp/qualy_quanty.tex
\subsection{Qualitative and Quantitative Results}

We assess the performance of our approach using both synthetic and real-world datasets. Firstly, we validate the effectiveness of our method on synthetic datasets, utilizing the first 20 frames as our observation set for training dynamic modeling capabilities. For future prediction, we employ the subsequent 10 frames, comparing the ground truth with our model's predictions of future frames. Additionally, we benchmark our approach against the most relevant methods in dynamic scene modeling and physics-informed learning, including PAC-NeRF \cite{li2023pacnerf}, Dynamic 3D Gaussians \cite{luiten2023dynamic}, and 4D Gaussian Splatting \cite{wu20234dgaussians}.

\input{sec/table/ft_pred.tex}

\input{sec/table/dy_recon.tex}

The qualitative results are presented in \cref{fig:qualitative_mpm}. We also report quantitative results by computing the Chamfer Distance (CD) and Earth Mover's Distance (EMD). In all tables, the Chamfer Distance is measured based on squared distance, with units expressed as $1 \times 10^3 mm^{2}$. The quantitative analysis of dynamic reconstruction, shown in \cref{tab:dy_recon}, reveals that \method can accurately simulate object dynamics for the synthetic data.

In \cref{tab:ft_pred}, we show our method's capability in predicting future frames. Our method outperforms PAC-NeRF across CD, EMD, Peak Signal-to-Noise Ratio (PSNR), and Structural Similarity Index (SSIM) metrics. We show qualitative results in \cref{fig:qualitative_mpm}. We observe that \method gives faithful reconstruction and future prediction that closely aligns with the ground truth, even though the simulation engines are different. This shows that \method is not only expressive but also allows efficient identification through gradient-based optimization.

%% file: sec/table/ft_pred.tex
\begin{table}[!t]
        \centering
        \setlength{\tabcolsep}{3pt}
        \begin{small}
                \begin{tabular}{c|l|ccccccc|c}
                        \shline
                        \multicolumn{2}{c|}{} & Torus                         & Cross          & Cream          & Apple          & Paste          & Chess          & Banana         & Mean                            \\
                        \shline

                        \multirow{2}*{\rotatebox{90}{{\tiny CD$\downarrow$}}}
                                              & \method (ours)                & \textbf{2.38}  & \textbf{1.57}  & 2.22           & \textbf{1.87}  & \textbf{7.03}  & \textbf{2.59}  & \textbf{18.48} & \textbf{5.16}  \\
                                              & PAC-NeRF \cite{li2023pacnerf} & 2.47           & 3.87           & \textbf{2.21}  & 4.69           & 37.70          & 8.20           & 66.43          & 17.94          \\

                        \hline
                        \multirow{2}*{\rotatebox{90}{{\tiny EMD$\downarrow$}}}
                                              & \method (ours)                & 0.087          & \textbf{0.051} & 0.094          & \textbf{0.076} & \textbf{0.126} & \textbf{0.095} & \textbf{0.135} & \textbf{0.095} \\
                                              & PAC-NeRF \cite{li2023pacnerf} & \textbf{0.055} & 0.111          & \textbf{0.083} & 0.108          & 0.192          & 0.155          & 0.234          & 0.134          \\

                        \hline
                        \multirow{2}*{\rotatebox{90}{{\tiny PSNR$\uparrow$}}}
                                              & \method (ours)                & 16.83          & \textbf{16.93} & \textbf{15.42} & \textbf{21.55} & \textbf{14.71} & \textbf{16.08} & \textbf{17.89} & \textbf{17.06} \\
                                              & PAC-NeRF \cite{li2023pacnerf} & \textbf{17.46} & 14.15          & 15.37          & 19.94          & 12.32          & 15.08          & 16.04          & 15.77          \\

                        \hline
                        \multirow{2}*{\rotatebox{90}{{\tiny SSIM$\uparrow$}}}
                                              & \method (ours)                & \textbf{0.919} & \textbf{0.940} & \textbf{0.862} & \textbf{0.902} & \textbf{0.872} & \textbf{0.881} & \textbf{0.904} & \textbf{0.897} \\
                                              & PAC-NeRF \cite{li2023pacnerf} & 0.913          & 0.906          & 0.858          & 0.878          & 0.819          & 0.848          & 0.866          & 0.870          \\

                        \shline
                \end{tabular}
        \end{small}
        \caption{\small \textbf{Quantitative results of future prediction on synthetic data.} \method excels in short-term future prediction. Meanwhile, since we separate appearance and dynamics modeling, \method also maintains good rendering quality.}
        \label{tab:ft_pred}
\end{table}

%% file: sec/table/dy_recon.tex
\begin{table}[!t]
        \centering
        \setlength{\tabcolsep}{2pt}
        \begin{small}
                \centering
                \begin{tabular}{c|l|ccccccc|c}
                        \shline
                        \multicolumn{2}{c|}{} & Torus                            & Cross          & Cream          & Apple          & Paste          & Chess          & Banana          & Mean                            \\
                        \shline
                        \multirow{4}*{\rotatebox{90}{CD$\downarrow$}}
                                              & \method (ours)                   & \textbf{0.17}  & \textbf{0.48}  & \textbf{0.36}  & \textbf{0.38}  & \textbf{0.19}  & 1.80            & \textbf{2.60}  & \textbf{0.85}  \\
                                              & PAC-NeRF \cite{li2023pacnerf}    & 4.92           & 1.10           & 0.77           & 1.11           & 3.14           & \textbf{0.96}   & 2.77           & 2.11           \\
                                              & Dy-Gaus \cite{luiten2023dynamic} & 579            & 773            & 479            & 727            & 2849           & 764             & 2963           & 1305           \\
                                              & 4D-Gaus \cite{wu20234dgaussians} & 11.12          & 1.77           & 2.87           & 2.23           & 1.95           & 3.97            & 7.13           & 4.43           \\
                        \hline
                        \multirow{4}*{\rotatebox{90}{EMD$\downarrow$}}
                                              & \method (ours)                   & \textbf{0.040} & \textbf{0.037} & \textbf{0.031} & \textbf{0.033} & \textbf{0.022} & 0.063           & \textbf{0.052} & \textbf{0.040} \\
                                              & PAC-NeRF \cite{li2023pacnerf}    & 0.056          & 0.052          & 0.041          & 0.045          & 0.054          & \textbf{0.052 } & 0.062          & 0.052          \\
                                              & Dy-Gaus \cite{luiten2023dynamic} & 0.857          & 0.955          & 0.783          & 0.903          & 1.739          & 0.985           & 1.591          & 1.116          \\
                                              & 4D-Gaus \cite{wu20234dgaussians} & 0.130          & 0.078          & 0.089          & 0.088          & 0.070          & 0.097           & 0.112          & 0.095          \\
                        \hline
                        \shline
                \end{tabular}
        \end{small}
        \caption{\small \textbf{Quantitative results of dynamic reconstruction on synthetic data.} \method has excellent geometric accuracy in dynamic reconstruction.}
        \label{tab:dy_recon}
\end{table}

%% file: sec/exp/real_sample.tex
\input{sec/img/real_results.tex}

% \subsection{Real World Samples}

We also evaluate \method on real-world samples. It is difficult for both NeRF \cite{mildenhall2020nerf} and 3D Gaussian Splatting \cite{kerbl3Dgaussians} to reconstruct the correct geometric information under extremely sparse camera views. However, due to 3D Gaussians' explicit representation that we can directly operate Gaussian kernels, we could reconstruct the static scenes and dynamic scenes in a different coordinate system and then align them using a registration network mentioned in \cref{sec:details}, which is hard to do under an implicit representation. We show the registration process in \cref{fig:regist}. Results on real-world samples can be found in \cref{fig:real_results} and \cref{fig:teaser}.

%% file: sec/img/real_results.tex
\begin{figure}[!t]
        \begin{center}
                \includegraphics[width=\linewidth]{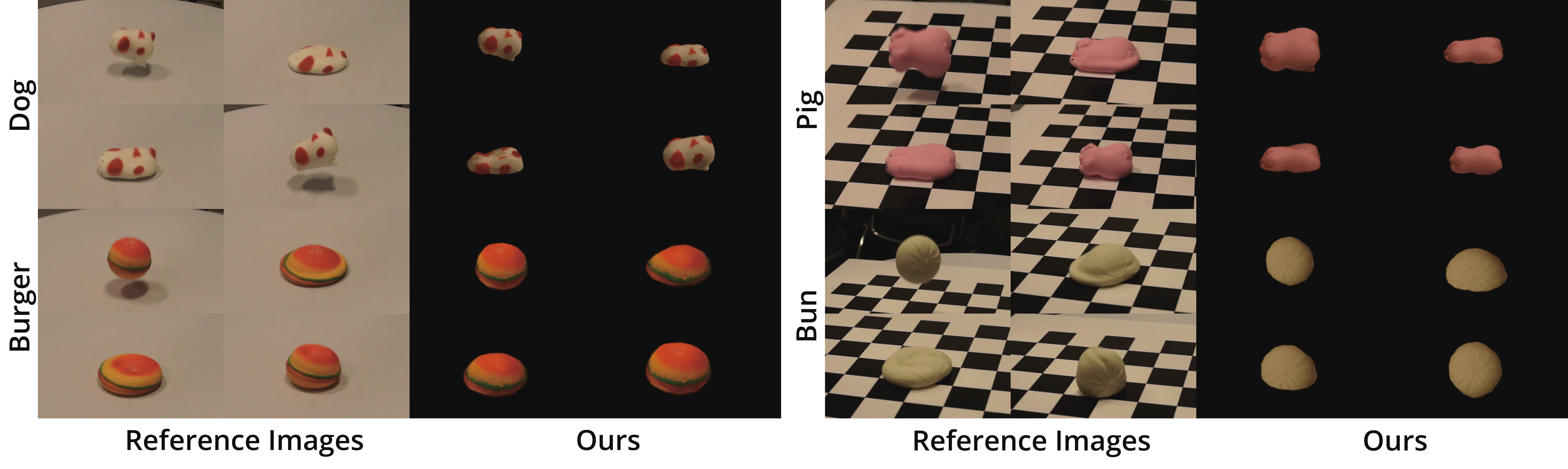}
        \end{center}
        \caption{\small \textbf{Qualitative results of future prediction on real-world samples.} Predicted dynamics closely follow real observations.}
        \label{fig:real_results}
\end{figure}

%% file: sec/exp/digital.tex
\subsection{Generalization to New Conditions}

In addition to future prediction, we also show that our method essentially creates a digital asset of the object from the multi-view videos, allowing dynamic simulation under different unseen environmental conditions. In \cref{fig:edit_bc}, we edit the boundary conditions, such as adjusting the positions and the stickiness of the ground plane. In \cref{fig:edit_env}, we edit physical conditions, initialization conditions, and environmental conditions, such as object properties (making them softer or harder), initial velocities, and environmental gravity. Please check our project website$^*$ for videos of the results for a more expressive illustration.
% : \projectURL.
\def\thefootnote{*}\footnotetext{\projectURL}

\input{sec/img/edit_bc.tex}
\input{sec/img/edit_env.tex}

%% file: sec/img/edit_bc.tex
\begin{figure}[!h]
        \centering
        \includegraphics[width=\textwidth]{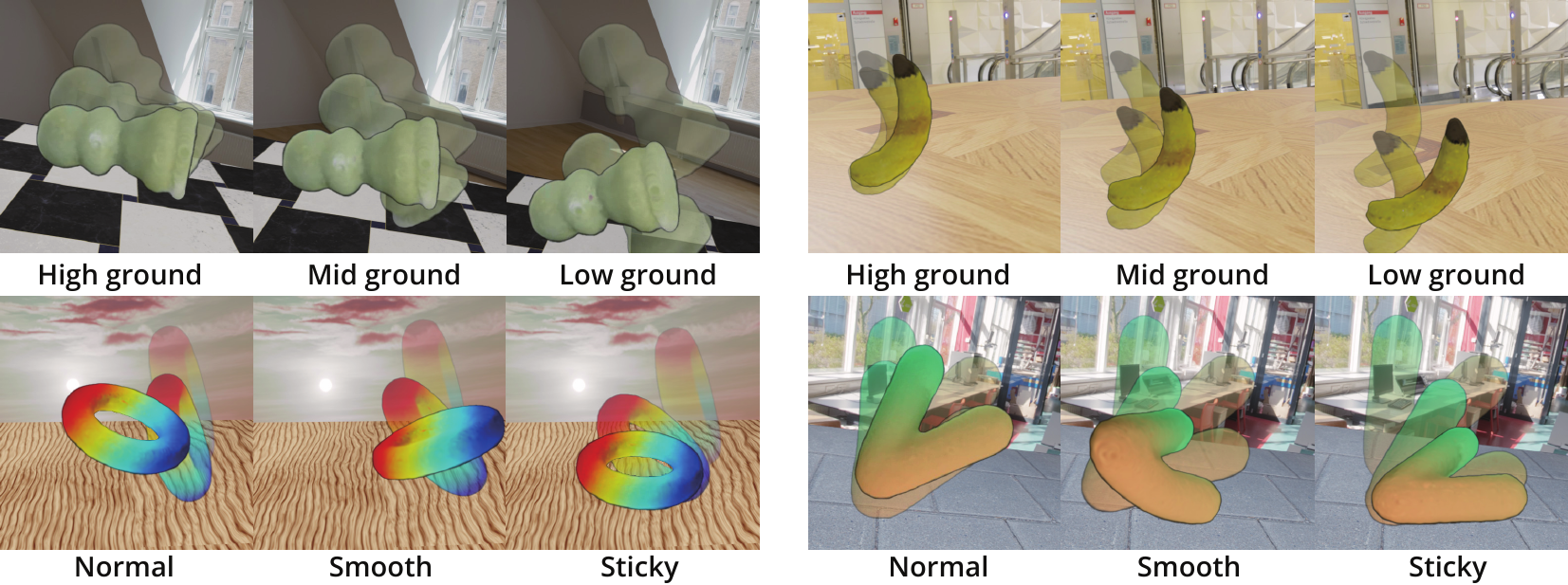}
        \caption{We edit boundary conditions in these demos. Changing grounds' position and using smooth or sticky ground.}
        \label{fig:edit_bc}
\end{figure}

%% file: sec/img/edit_env.tex
\begin{figure}[!h]
        \centering
        \includegraphics[width=\textwidth]{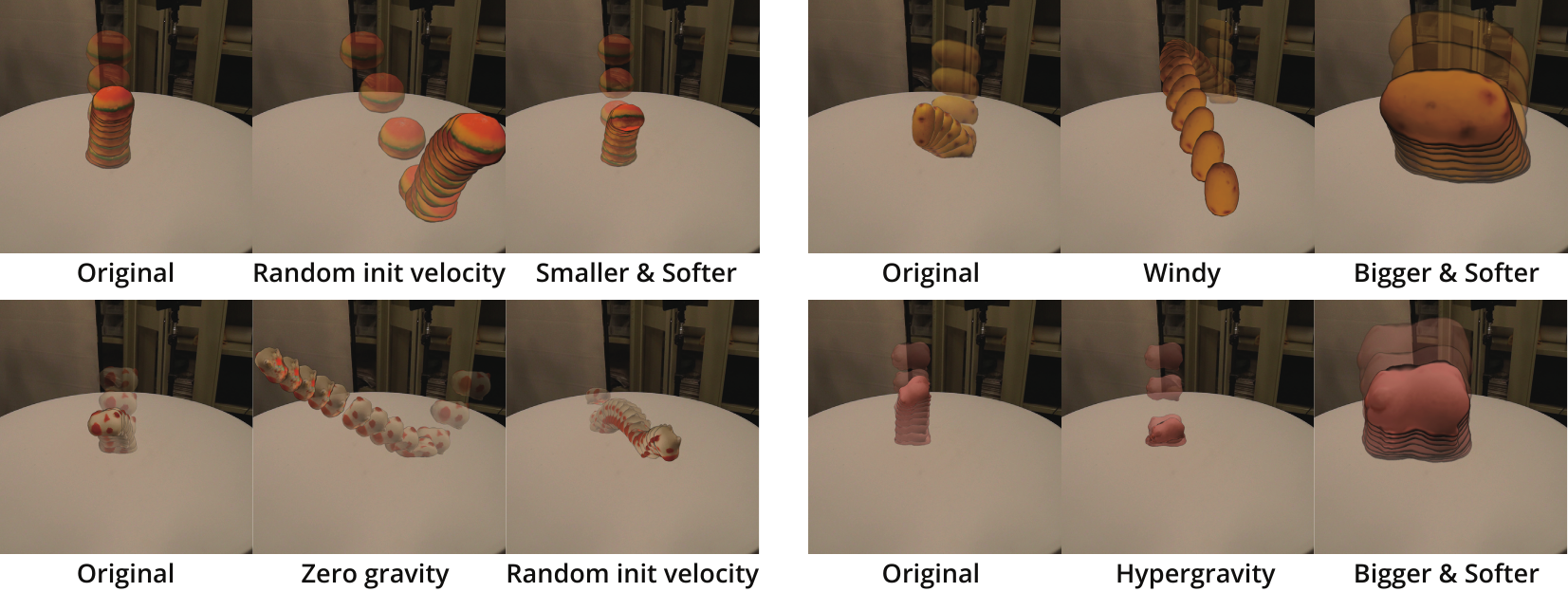}
        \caption{We edit physical conditions, initial velocities and gravities in these demos.}
        \label{fig:edit_env}
\end{figure}

%% file: sec/7_conclusion.tex
\section{Conclusion}

In this paper, we introduce \method, a novel framework designed to acquire simulatable digital assets of elastic objects from video observations. By integrating a 3D Spring-Mass model, \method allows the reconstruction of object appearance and dynamics. A key feature of our approach is the distinct separation between the learning processes for appearance and physics, thereby circumventing potential issues with optimization interference.
% The abstract modeling approach of \method, coupled with its modest requirements on particle distribution and count, endows it with a good capacity for dynamic modeling. 
We evaluate \method on both synthetic and real-world datasets, demonstrating its capability to reconstruct geometry, appearance, and physical dynamic properties.
Moreover, our method demonstrates improved capabilities by predicting short-term future dynamics under different environmental conditions. This showcases its strength in identifying physical properties from observational data and predicting the dynamics of reconstructed digital assets.

% \myparagraph{Limitations.}

%% file: sec/supplementary.tex
\section{Data flow}
\begin{figure*}[hb]
    \centering
    \includegraphics[width=0.9\linewidth]{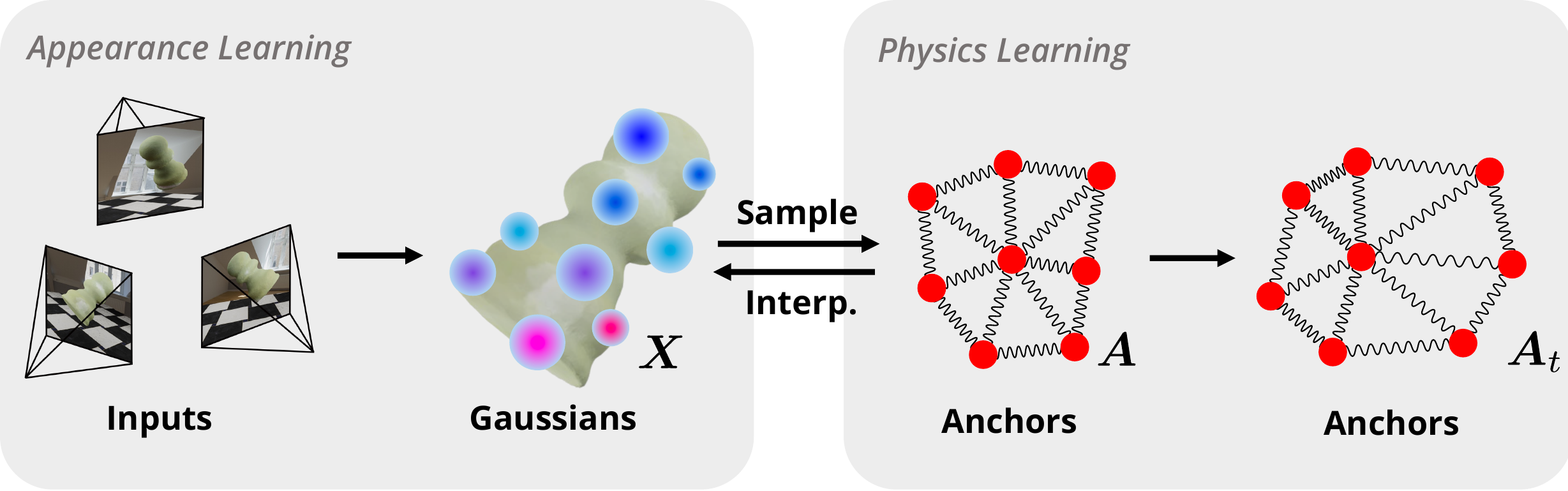}
    \caption{Simpler layout of the pipeline. In appearance learning, we optimize Gaussian kernels for static image rendering. We sample anchors $\boldsymbol{A}$ from Gaussians' center $\boldsymbol{X}$. In physics learning, we optimize physical parameters among anchors $\boldsymbol{A}$. After simulation, at time $t$, the updated Gaussians' center $\boldsymbol{X}_t$ could be interpolated from $\boldsymbol{A}_t$.}
    \label{fig:data_flow}
\end{figure*}

To clarify the decoupling of appearance learning and physical learning, we provide a simpler layour of our pipeline, as shown in \cref{fig:data_flow}.

\section{More Implementation Details}
We use a single NVIDIA RTX 3090 GPU for reconstruction. For static scene reconstruction, we follow the configuration prescribed by 3D Gaussian Splatting \cite{kerbl3Dgaussians}, which takes approximately 10 minutes to optimize for the synthetic data. The dynamic reconstruction takes 300 iterations. We employ 2048 anchor points for our Spring-Mass model, with each anchor linked to $n_k=256$ neighbors through springs. We set $n_b=16$ and $n_c=16$. For each sequence, we assume mass $m_0=1$ and damping factor $\zeta_0=0.1$. The weighting coefficient for the D-SSIM term $\lambda_{\mathit{d-ssim}}$ is set to $0.2$ for static reconstruction and $0.05$ for dynamic reconstruction. In our experiments, we use a nonlinear spring force, setting $p_k=0.5$, and for Inverse Distance Weighting (IDW) interpolation, we arbitrarily choose $p_b=0.5$.

Following the practice from PAC-NeRF \cite{li2023pacnerf}, we first independently optimize the initial velocity vector $\boldsymbol{v}_0$, utilizing only a few frames captured before the object interacts with the environment.

In terms of the Gaussian kernels' parameters, we optimize all of them during static scene reconstruction while maintaining a constant scaling scalar $s_0$ for all kernels. We have found that uniform scaling across all kernels in static scene reconstruction results in a more evenly distributed point cloud and anchor points. This consistency markedly improves the dynamic model's simulation capabilities by making the kernels' spatial distribution more uniform. It is important to note that we do \emph{not} preserve the scaling scalar as constant $s_0$ during this refinement phase. Instead, we assign a unique scaling scalar $s_i$ to each Gaussian kernel associated with each anchor point.

\section{Ablation Study}

\input{sec/img/ablation.tex}
\input{sec/table/ablation.tex}

In our approach, we employ a soft vector $\eta$ to dynamically regulate both the quantity and intensity of springs linked to the anchor points. This strategy is illustrated in \cref{tab:ablation}, showcasing its effectiveness. Our method's capability to simulate using very sparse anchors allows for the individual optimization of physical parameters for each anchor point. This contrasts with PAC-NeRF, which utilizes tens of thousands of particles, making it challenging to optimize the physical parameters for each particle infeasible. Consequently, PAC-NeRF faces limitations in accurately modeling objects composed of heterogeneous materials. In contrast, our methodology is adept at handling such complexities. As depicted in \cref{fig:ablation} and \cref{tab:ablation}, we present the outcomes on a heterogeneous object that is segmented into various sections, each with distinct physical properties, thereby demonstrating our model's superior adaptability in capturing the nuanced dynamics of objects with variable material composition.

\section{Limitations and Future Work}
Currently, \method is constrained to modeling elastic objects due to fixed spring lengths in our formulation; these lengths are constants established at the onset of dynamic simulation. Future work should aim to incorporate plastic deformation into the framework. This would involve developing a method to dynamically adjust the original lengths of the springs, also can make and break spring relationships during the simulation, allowing for the accurate modeling of materials that exhibit both elastic and plastic behavior.

Besides, our method focuses on simulating a single object colliding with the ground surface, while multi-object interaction is a fascinating topic but requires new model design (e.g., establishing new springs) and a more thorough evaluation under various challenging scenarios, which we identify as an excellent direction to expand our method. Other directions include considering more complicated boundary conditions and external actions.
% While the reconstructed digital asset can be of great value for immersive and robotic applications, a more thorough investigation is necessary to account for more realistic, in-the-wild scenarios and more complicated physical interactions.

Lastly, for better evaluation, a more comprehensive real world dataset with high spatial and temporal resolutions should be collected, including more diverse objects, materials, and interactions. This dataset should also include more challenging scenarios, such as occlusions, lighting changes, and camera motion. This kind of dataset will help to evaluate the robustness and generalization of related methods.

%% file: sec/img/ablation.tex
\begin{figure}[!t]
        \begin{center}
                \includegraphics[width=\linewidth]{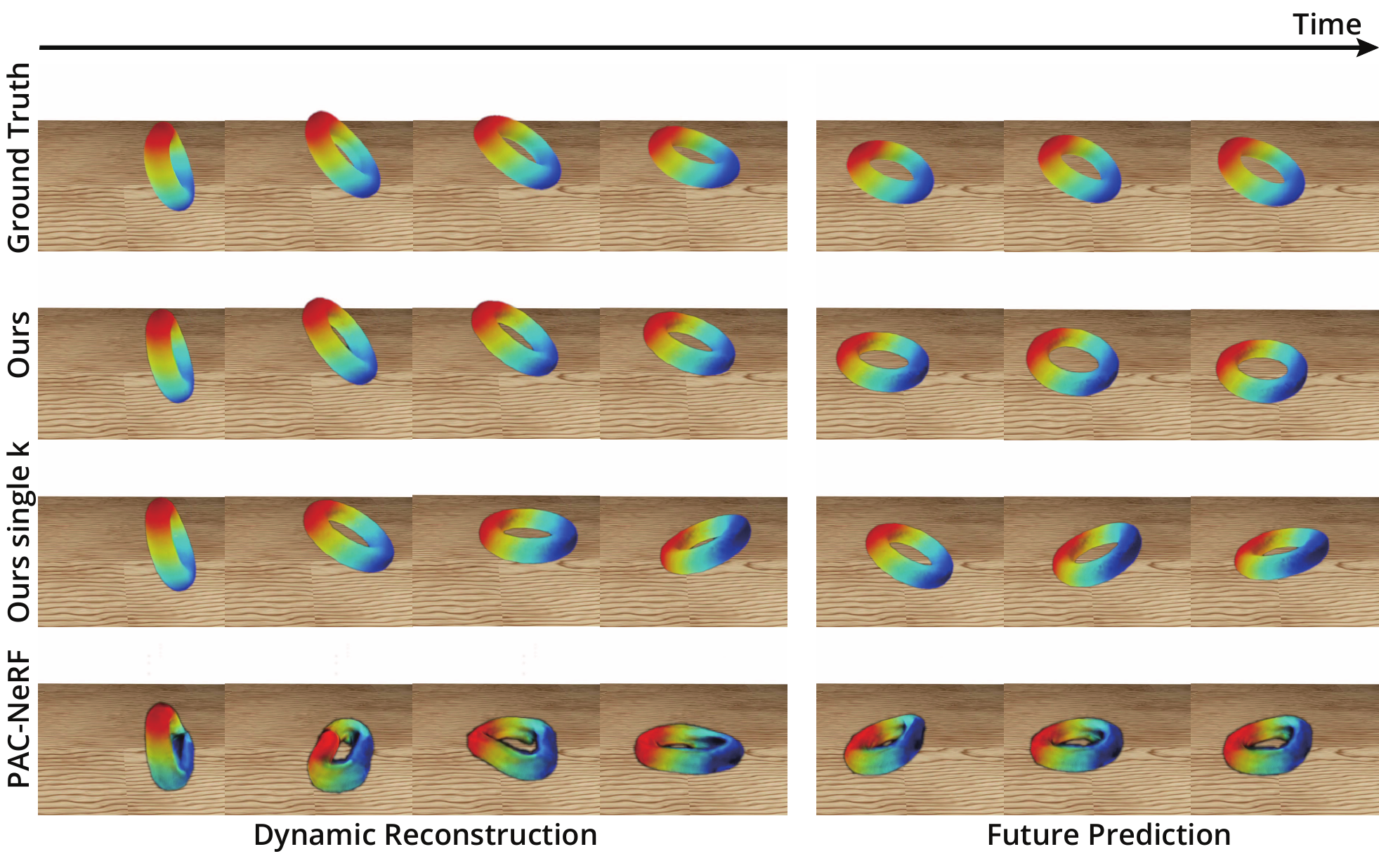}
        \end{center}
        \caption{Ablation study of the effectiveness of optimizing physical parameters for each particle rather than optimizing a single global parameter, on a heterogeneity object. The results shows that optimizing a single global parameter is not able to accurately model objects with complex physical properties.}
        \label{fig:ablation}
\end{figure}

%% file: sec/table/ablation.tex
\begin{table}[!h]
        \centering
        \setlength{\tabcolsep}{3pt}
        \begin{small}
                \begin{tabular}{lcccccc}
                        \toprule
                                                       & \multicolumn{3}{c}{Dynamic Reconstruction} & \multicolumn{3}{c}{Future Prediction}                                                                     \\
                        \cmidrule(lr){2-4}\cmidrule(lr){5-7}
                                                       & CD$\downarrow$                             & PSNR$\uparrow$                        & SSIM$\uparrow$ & CD$\downarrow$ & PSNR$\uparrow$ & SSIM$\uparrow$ \\
                        \midrule
                        \method (ours)                 & \textbf{0.18}                              & \textbf{27.08}                        & \textbf{0.967} & \textbf{2.04}  & \textbf{17.63} & \textbf{0.927} \\
                        \method w/o soft vector $\eta$ & 0.56                                       & 25.36                                 & 0.959          & 13.28          & 13.91          & 0.881          \\
                        \method, single $k$            & 3.22                                       & 23.02                                 & 0.940          & 6.56           & 14.45          & 0.892          \\
                        PAC-NeRF \cite{li2023pacnerf}  & 8.66                                       & 19.87                                 & 0.916          & 5.70           & 15.65          & 0.894          \\
                        \bottomrule
                \end{tabular}
        \end{small}
        \caption{\small Ablation study. We demonstrate the importance of optimizing parameters for each anchor point individually as well as using a soft vector $\eta$. Optimizing parameters for each anchor point allows \method to have a higher degree of freedom in modeling physics, and the soft vector $\eta$ gives a more flexible formulation.}
        \label{tab:ablation}
\end{table}